\documentclass[lettersize,journal]{IEEEtran}
\usepackage{amsmath,amsfonts}
\usepackage{algorithmic}
\usepackage{algorithm}
\usepackage{array}
\usepackage[caption=false,font=normalsize,labelfont=sf,textfont=sf]{subfig}
\usepackage{textcomp}
\usepackage{stfloats}
\usepackage{url}
\usepackage{verbatim}
\usepackage{graphicx}
\usepackage{cite}
\usepackage{booktabs}
\usepackage[colorlinks,linkcolor=blue,citecolor=blue]{hyperref}

\begin{document}

\title{Low-Dimensional Representation-Driven TSK Fuzzy System for Feature Selection}

\author{Qiong Liu, Mingjie Cai, Qingguo Li
        % <-this % stops a space
\thanks{This work is supported by NSFC (No. 12471431, 12231007), Hunan Provincial Natural Science Foundation of China (No. 2023JJ30113) and Guangdong Basic and Applied Basic Research Foundation (No. 2023A1515012342). (Corresponding authors: Mingjie Cai.)}% <-this % stops a space
\thanks{Qiong Liu, Mingjie Cai and Qingguo Li are with the College of Mathematics, Hunan University, Changsha 410082, China (e-mail: qliu6666@163.com; cmjlong@163.com; liqingguoli@aliyun.com).}
}
% \author{*
% \thanks{*}% <-this % stops a space
% % \thanks{Manuscript received April 19, 2021; revised August 16, 2021.}
% }

\markboth{*}
{Shell \MakeLowercase{\textit{et al.}}: A Sample Article Using IEEEtran.cls for IEEE Journals}
%\IEEEpubidadjcol 文字附近插入使得版权字与正文文字不重叠
%\eqref{deqn_ex1a} 用于引用公式

\maketitle

\begin{abstract}
Feature selection can select important features to address dimensional curses. Subspace learning, a widely used dimensionality reduction method, can project the original data into a low-dimensional space. However, the low-dimensional representation is often transformed back into the original space, resulting in information loss. Additionally, gate function-based methods in Takagi-Sugeno-Kang fuzzy system (TSK-FS) are commonly less discrimination. To address these issues, this paper proposes a novel feature selection method that integrates subspace learning with TSK-FS. Specifically, a projection matrix is used to fit the intrinsic low-dimensional representation. Subsequently, the low-dimensional representation is fed to TSK-FS to measure its availability. The firing strength is slacked so that TSK-FS is not limited by numerical underflow. Finally, the $\ell _{2,1}$-norm is introduced to select significant features and the connection to related works is discussed. The proposed method is evaluated against six state-of-the-art methods on eighteen datasets, and the results demonstrate the superiority of the proposed method.
\end{abstract}


\section{INTRODUCTION}
\IEEEPARstart{O}{ver} the past decade, stored data have been generated at an unprecedented rate, but the computational efficiency has not been able to satisfy the gradually increasing demand. There is a large amount of redundance and noise in the data, which limits the performance of machine learning algorithms and increases the computational complexity. Dimensionality reduction can reduce the dimension of data to improve the performance of machine learning algorithms and decrease the computational complexity. Therefore, dimensionality reduction has been widely used as a crucial preprocessing step in various fields, such as computer vision~\cite{he2016deep,zhu2020unsupervised}, natural language processing~\cite{hirschberg2015advances,sidorov2014syntactic}, and bioinformatics~\cite{liu2019biogenesis,zhang2024self}.

Dimensionality reduction can generally be divided into two categories: feature extraction and feature selection~\cite{krizhevsky2012imagenet,zou2024multi}. Feature extraction apply a linear or non-linear transformation to original data, generating a new data form. For example, Principal Component Analysis (PCA) is a well-known linear feature extraction method, which can project the original data into a low-dimensional space by maximizing the variance of the projected data~\cite{abdi2010principal}. Subsequently, Nie et al. proposed a row-sparse PCA that is suitable for various data structures, and an efficient coordinate descent method to solve the proposed optimization problem~\cite{nie2024row}. The variational auto-encoder serves as a non-linear feature extraction method that learns the probability distribution of a low-dimensional representation to generate data through random sampling~\cite{kingma2013auto}. Afterwards, Caterini et al. proposed the Hamiltonian variational auto-encoder, which yields a low-variance unbiased estimator of evidence lower bounds and its gradient, utilizing the reparameterization trick~\cite{caterini2018hamiltonian}. However, feature extraction methods are confronted with the problem that the importance of original features to the data is not recognized. For instance, feature extraction may not be a splendid choice if the significance of a gene segment for a certain disease needs to be evaluated. 

Feature selection methods select the most representative features from the original data. It mainly includes group-based score methods and single-feature rank-based methods. The former employs a heuristic search algorithm to select a group of features, leading to a locally optimal solution and high computational complexity. In references~\cite{wan2021interactive,dai2024feature,wang2022feature}, they constructed the variant of information entropy as an evaluation function to find the optimal feature subset. In theory, the number of searches is $\frac{m(m+1)}{2}$ in the worst case and the parallel computing technology is not available.

The single-feature rank-based method considers the importance of single feature, resulting in faster computational efficiency and a satisfactory result. Graph embedding-based methods~\cite{wang2016sparse,wang2022joint,wei2012graph} and trace ratio-based methods~\cite{wen2018robust,zhao2018trace,li2023sparse} mostly leveraged $\ell _{2,p}$-norm and $\ell _{p}$-norm, $p=\{0,1,2\}$, to select the significant features, taking into consideration the performance, efficiency and interpretability. Similarly, subspace learning-based methods~\cite{wang2015subspace,nie2020subspace,shang2020sparse} utilized these norms to select the significant features. They aim to model the data according to different conditions or assumptions. However, when faced with complex data, the performance of these methods may not be satisfactory. To discover intrinsic relations between feature and label, causal graph-based methods is proposed~\cite{yu2019multi}. It tested the conditional independence between features and features as well as between features and labels, then selected a feature subset based on Markov Blanket~\cite{yu2018mining,aliferis2010local}. However, the construction of a causal graph has high computational complexity. In 2018, Zheng et al. proposed a method, transforming the combinatorial optimization problem to the continuous optimization problem to construct the causal graph~\cite{zheng2018dags}. Nevertheless, they lack the ability to handle large-scale or high-dimensional data.

Takagi–Sugeno–Kang fuzzy system (TSK-FS), a nonlinear model, has been widely used in various fields~\cite{wu2019functional,cui2022layer}. It offers high interpretability based on fuzzy rule base. Therefore, many feature selection methods based on TSK-FS have been proposed~\cite{ji2024convergence}. Gate function-based methods constrain the antecedent parameter or weight the consequent parameter to select the significant features. Chen et al. leveraged the gate function to control the membership value, selecting the significant features~\cite{chen2011integrated}. Chakraborty et al. proposed a neuro-fuzzy scheme for feature selection and classification, using four-layer feed-forward neural network~\cite{chakraborty2004neuro}. On the other hand, Gong et al. proposed a refined softmin to alleviate numerical underflow, and the group lasso loss function to select the significant features~\cite{gong2024embedded}. These methods have trained a model using all features, resulting in high computational complexity and weak performance. In addition, many scholars have focused on the problem of numerical underflow in TSK-FS, but it is still a challenge with available computer technology. 

Our contributions are summarized as follows:
\begin{enumerate}
        \item The low-dimensional representation generated by the projection matrix is no longer transformed back into the original space, but its validity is ensured by classification.
        \item The low-dimensional representation is used to train TSK-FS, decreasing the interference of redundant or noisy features and the computational complexity. Meanwhile, the firing strength is directly learned, which avoids learning antecedent parameters and numerical underflow.
        \item The proposed optimization problem is solved by the alternative optimization strategy. Experiments on eighteen datasets show the effectiveness and superiority of the proposed method.
\end{enumerate}
The rest of this paper is organized as follows. Section~\ref{sec:pre} introduces the preliminaries. Section~\ref{sec:method} presents the proposed method. Section~\ref{sec:exp} shows the experimental results. Section~\ref{sec:exp} concludes this paper.
\section{PRELIMINARIES}\label{sec:pre}
\subsection{TSK Fuzzy System}
Given an input data $\mathbf{X}=\left[\mathbf{x}_1, \dots, \mathbf{x}_n\right]^{\mathrm{T}}\in \mathbb{R}^{n\times m}$ and label $\mathbf{y}\in \mathbb{R}^{n\times 1}$, where $n$ and $m$ are the number of samples and features respectively, the boldfaced majuscule denotes a matrix, the boldfaced minuscule denotes a vector. TSK-FS is a high interpretability model, benefiting from "IF-THEN" rules that can be formulated as follows:
\begin{align*}
        &\text{IF} \quad x_1 \text{ is } A_{r,1} \text{ and } x_2 \text{ is } A_{r,2} \text{ and } \cdots \text{ and } x_d \text{ is } A_{r,d},\\
        &\text{THEN} \quad o_r(x) = p_{r,0} + \sum_{i=1}^{d} p_{r,i} x_i,\\
        &r=1,2,\cdots,k,
\end{align*}
where $x_i$ represents the $i$-th feature of the sample $x$, $A_{r,i}$ is a fuzzy set of the rule $r$ w.r.t. the $i$-th feature, and $p_{r,i}$ denotes the consequent parameter. IF $\forall i>0, p_{r,i}=0$, we call the model as zero-order TSK-FS. In general, the fuzzy set $A_{r,i}$ is modeled by the Gaussian function, which can be represented as:
\begin{equation}\label{eq:gaussian}
        \mu _{r,i}(x) = e^{\left(-\frac{\left\|x_i - m_{r,i}\right\|_2^2}{2\delta _{r,i}^2}\right)},
\end{equation} 
where $m_{r,i}$ and $\delta _{r,i}$ are the center and spread, respectively. The $m_{r,i}$ and $\delta _{r,i}$ are also called the antecedent parameters. They are often initialized by fuzzy $c$-means clustering. To fuse the membership degree of each rule, T-norm is used to compute the firing strength. A binary operation $\mathcal{T}$: $[0,1]\times [0,1]\rightarrow [0,1]$ is T-norm if it satisfies the following properties:
\begin{enumerate}
        \item Commutativity: $\mathcal{T}(a,b)=\mathcal{T}(b,a)$,
        \item Associativity: $\mathcal{T}(a,\mathcal{T}(b,c))=\mathcal{T}(\mathcal{T}(a,b),c)$,
        \item Monotonicity: $a\leq b$ and $c\leq d$ implie $\mathcal{T}(a,c)\leq \mathcal{T}(b,d)$,
        \item Identity element: $\mathcal{T}(a,1)=a$,
        \item Zero element: $\mathcal{T}(a,0)=0$.
\end{enumerate}
Generally, the product T-norm and minimum T-norm are commonly used in TSK-FS. The former may result in numerical underflow. The latter is not differentiable, leading to the difficulty of optimization. Two ways to compute the firing strength are listed as follows:
\begin{align}
        &\text{Product T-norm: } f_r(x) = \prod_{i=1}^{d} \mu _{r,i}(x),\\
        &\text{Minimum T-norm: } f_r(x) = \min_{i=1}^{d} \mu _{r,i}(x).
\end{align}
The firing strength is commonly normalized to ensure that the value lies within the range of $[0,1]$. The normalized form of firing strength is listed as follows:
\begin{eqnarray}
        \overline{f}_r(x) = \frac{f_r(x)}{\sum_i f_i(x)}.
\end{eqnarray}
Then, the output of TSK-FS can be represented as:
\begin{equation}
        \hat{y}(x) = \sum_{r=1}^{k} \overline{f}_r(x) o_r(x).
\end{equation}
\subsection{Solution}
The antecedent parameters and consequent parameters are learned by the least squares method. The objective function can be represented as:
\begin{equation}\label{eq:obj1}
        \min _{p_{r,i}, m_{r,i}, \delta _{r,i}} \sum_{i=1}^n (\hat{y}(x_i) - y_i)^2.
\end{equation}
Due to the non-convexity of the objective function, the optimization problem is intractable. A common approach is to use gradient descent to solve Eq. \eqref{eq:obj1}. Their gradients are listed as follows:
\begin{align}
        &\nabla m_{r,d} = 2\sum_{i=1}^n \hat{y}(x_i)(\hat{y}(x_i) - y_i)(1-\overline{f}_r(x_i))\frac{x_i - m_{r,d}}{\delta ^2},\\
        &\nabla \delta _{r,d} = 2\sum_{i=1}^n \hat{y}(x_i)(\hat{y}(x_i) - y_i)(1-\overline{f}_r(x_i))\frac{(x_i - m_{r,d})^2}{\delta ^3},\\
        &\nabla p_{r,d} = 2\sum_{i=1}^n (\hat{y}(x_i) - y_i)\overline{f}_r(x_i)x_{i,d},\\
        &\nabla p_{r,0} = 2\sum_{i=1}^n (\hat{y}(x_i) - y_i)\overline{f}_r(x_i).
\end{align}
Then, the parameters are updated as follows:
\begin{equation}
        \tau^{(t+1)} = \tau^{(t)} - \eta \nabla \tau^{(t)},
\end{equation}
where $\tau$ denotes these parameters and $\eta >0$ is the learning rate.
\section{The Proposed Method}\label{sec:method}
\begin{figure*}[htbp!]
        \centering
        \includegraphics[width=0.6\textwidth]{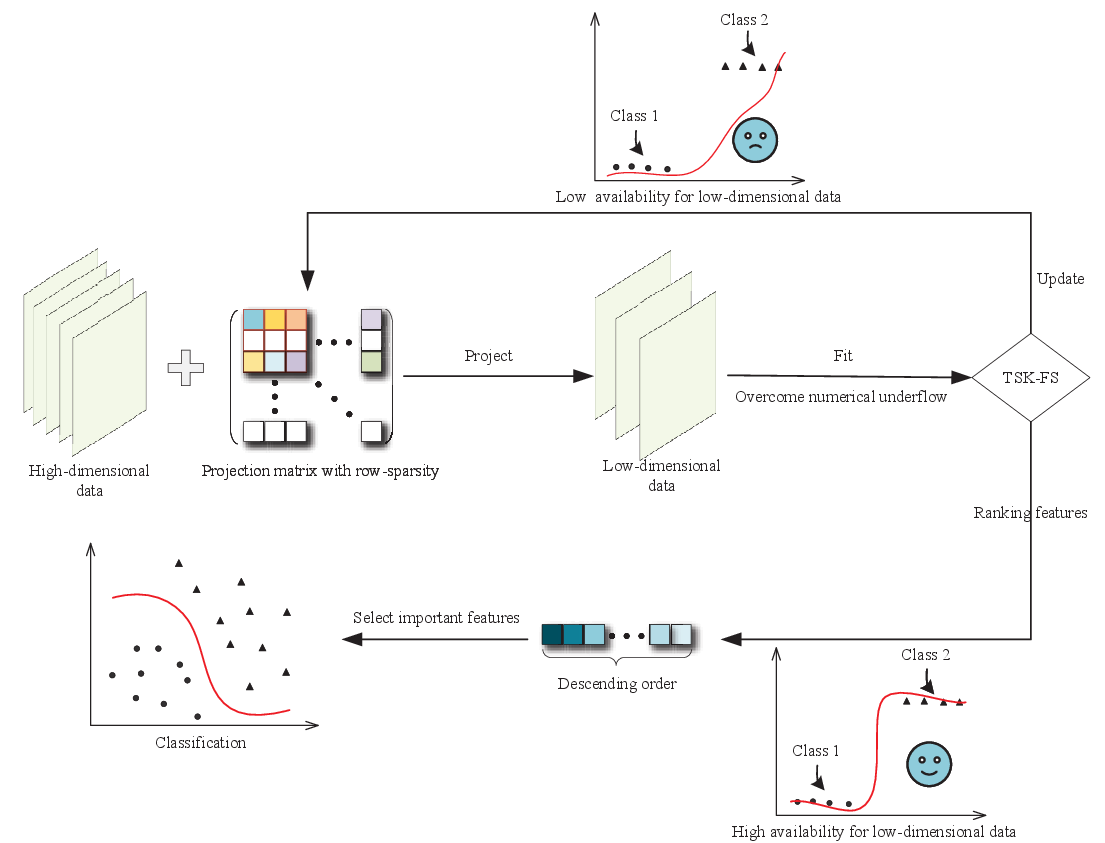}
        \caption{The schematic of the proposed method.}
        \label{fig:framework}
\end{figure*}

Feature selection can effectively reduce the dimension of data to improve the performance and decrease the computational complexity of a model in subsequent tasks. For feature selection, the high performance and interpretability is often required. TSK-FS offers interpretability through fuzzy inference based on fuzzy rule base. Nevertheless, a lot of fuzzy rules is generated, leading to the high computational complexity and numerical underflow. In addition, the selected features by the gate function are less discriminative. Based on this, we propose a novel feature selection method that combines subspace learning to obtain more discriminative features. The schematic of our method is shown in Fig.~\ref{fig:framework}.
\subsection{Low-dimensional Representation Embedding}
Commonly, a TSK-FS is successful if its classification performance is excellent for data. Likewise, we expect that TSK-FS can classify the low-dimensional representation of data well, so that the quality of low-dimensional representation is credible.

Let $\mathbf{\hat{X}}\in \mathbb{R}^{n\times d}$ be the low-dimensional representation of a data, where $n$ is the number of samples and $d$ is the dimension of the low-dimensional representation. Then, the low-dimensional TSK-FS with $k$ rules can be formulated as:
\begin{align*}
        &\text{IF} \quad \hat{x}_1 \text{ is } A_{r,1} \text{ and } \hat{x}_2 \text{ is } A_{r,2} \text{ and } \cdots \text{ and } \hat{x}_d \text{ is } A_{r,d},\\
        &\text{THEN} \quad y_r(\mathbf{x}) = p_{r,0} + \sum_{i=1}^{d} p_{i,r} \hat{x}_i,\\
        &r=1,2,\cdots,k.
\end{align*}
Note that we do not transform the label of data to one-hot encoding although one-hot encoding can improve the performance of model for classification to some extent~\cite{2021_pro1}. The number of labels affects the computational efficiency when one-hot encoding is employed. It should be noted that the final purpose is not to fit labels, but feature selection. Thus, a minor loss in performance is acceptable with great improvement in computational efficiency.

On the other hand, TSK-FS consists of two categories of parameters: antecedent parameters and consequent parameters. Antecedent parameters are used to compute the membership of a rule w.r.t. the input variable. Then, the firing strength is computed by T-norm. However, the usefulness of firing strength is influenced by the noise in data, kernel function, and numerical underflow. Therefore, we forgo learning the antecedent parameters and instead learn the firing strength directly. Although the fuzzy set of a rule w.r.t. the input variable is unknown, we still can recognize which of rules is more important. In addition, the IF clause can be recovered after training is completed if necessary. In other words, there is little to lose for TSK-FS, but additional parameters w.r.t. kernel function need not be considered, and numerical underflow is avoided. 

The firing strengths of rules w.r.t. samples are represented as the matrix $\mathbf{F}^{n\times k}$. Then, the objective function for the low-dimensional TSK-FS can be represented as:
\begin{equation}\label{eq:obj2}
        \min_{\mathbf{\hat{X}}, \mathbf{F}, \mathbf{P}, \mathbf{p}_0} \left\| (\mathbf{\hat{X}}\mathbf{P})\odot \mathbf{F}\mathbf{1}_k+\mathbf{F}\mathbf{p}_0-\mathbf{y} \right\|_2^2, \text{ s.t. } \mathbf{F}>0,
\end{equation}
where $\mathbf{P}\in \mathbb{R}^{d\times k}$ and $\mathbf{p}_0\in \mathbb{R}^{k\times 1}$. They are consequent parameters.

Eq. \eqref{eq:obj2} can ensure the availability of the low-dimensional representation in terms of classification. Additionally, the low-dimensional representation should be the intrinsic representation of the original data. Therefore, we introduce a projection matrix $\mathbf{Q}\in \mathbb{R}^{m\times d}$, which can transform the original data to the low-dimensional representation. In addition, $\ell _{2,1}$-norm is introduced to ensure the row sparsity of the projection matrix, selecting the discriminative features. The final objective function can be represented as:
\begin{equation}\label{eq:obj}
        \begin{split}
                &\min_{\mathbf{\hat{X}}, \mathbf{F}, \mathbf{P}, \mathbf{Q}, \mathbf{p}_0} \left\| (\mathbf{\hat{X}}\mathbf{P})\odot \mathbf{F}\mathbf{1}_k+\mathbf{F}\mathbf{p}_0-\mathbf{y} \right\|_2^2 + \gamma \left\| P\right\|_\mathrm{F}^2 \\
        &\alpha \left\| \mathbf{X}\mathbf{Q}-\mathbf{\hat{X}} \right\|_\mathrm{F}^2 + \beta \left\| \mathbf{Q} \right\|_{2,1}, \\
        &\text{ s.t. } \mathbf{F}>0, \mathbf{Q}^\mathrm{T}\mathbf{Q}=\mathbf{I},  
        \end{split}
\end{equation}
where $\alpha$, $\beta$ and $\gamma$ are hyperparameters. The consequent parameter $\mathbf{P}$ is regularized by the Frobenius norm to avoid overfitting.
\subsection{Solution}
Although the objective function is convex w.r.t. a specific variable, the optimization problem is intractable. We use the alternative optimization strategy to update variables.

For $\mathbf{\hat{X}}$, $\mathbf{F}$ and $\mathbf{P}$, adaptive moment estimation (Adam) method~\cite{kingma2014adam} is used to update them. Comparison with stochastic gradient descent, Adam can adaptively adjust the learning rate for each parameter. The update rule for $\mathbf{\hat{X}}$, $\mathbf{P}$ and $\mathbf{F}$ can be represented as:
\begin{equation}\label{eq:upx}
        \mathbf{\hat{X}}^{(t+1)} = \mathbf{\hat{X}}^{(t)} - \frac{\eta \hat{m}_t}{\sqrt{\hat{v}_t} + \epsilon},
\end{equation}
\begin{equation}\label{eq:upp}
        \mathbf{\hat{P}}^{(t+1)} = \mathbf{\hat{P}}^{(t)} - \frac{\eta \hat{m}_t}{\sqrt{\hat{v}_t} + \epsilon},
\end{equation}
\begin{equation}\label{eq:upf}
        \mathbf{\hat{F}}^{(t+1)} = \mathbf{\hat{F}}^{(t)} - \frac{\eta \hat{m}_t}{\sqrt{\hat{v}_t} + \epsilon},
\end{equation}
where $\eta$ and $\epsilon$ are the learning rate and a small constant, respectively. $\hat{m}_t$ and $\hat{v}_t$ are the bias-corrected first-order and second-order moment estimation of the gradient, respectively, where 
\begin{equation}
        \hat{m}_t = \frac{m_t}{1-\beta_1^t},
\end{equation}
\begin{equation}
        \hat{v}_t = \frac{v_t}{1-\beta_2^t}.
\end{equation}
The first-order and second-order moment estimation of the gradient, $m_t$ and $v_t$, are updated as follows:
\begin{equation}
        m_t = \beta_1 m_{t-1} + (1-\beta_1) \varphi ^{(t)},
\end{equation}
\begin{equation}
        v_t = \beta_2 v_{t-1} + (1-\beta_2) (\varphi ^{(t)})^2,
\end{equation}
where $\varphi ^{(t)}$ is the gradient of the objective function w.r.t. the corresponding variable at the $t$-th iteration. $\beta_1$ and $\beta_2$ are the decay rates of the first-order and second-order moment estimation, respectively. In general, $\beta_1 = 0.9$ and $\beta_2 = 0.999$. Then, the gradients of $\hat{X}_{ij}$ and $P_{ij}$ are listed as follows:
\begin{equation}
        \begin{split}
                \nabla \hat{X}_{ij} =& 2(\langle \mathbf{\hat{x}}^i\mathbf{P}, \mathbf{f}^i \rangle + \langle \mathbf{f}^i, \mathbf{p}_0 \rangle - y_i)(\langle \mathbf{f}^i, \mathbf{p}^j \rangle) + \\
                &2\alpha(\hat{X}_{ij}- \mathbf{x}^i\mathbf{q}_j), 
        \end{split}  
\end{equation}
\begin{equation}
        \begin{split}
                \nabla P_{ij} = &2\sum _{h=1}^n(\langle \mathbf{\hat{x}}^i\mathbf{P}, \mathbf{f}^i \rangle + \langle \mathbf{f}^i, \mathbf{p}_0 \rangle - y_i)(\hat{X}_{hi}F_{hj}) + \\
        &2\gamma P_{ij},
        \end{split}
\end{equation}
where $\langle \bullet \rangle$ denotes the inner product, $\mathbf{\hat{x}}^i$ is the $i$-th row of $\mathbf{\hat{X}}$, and $\mathbf{q}_j$ is the $j$-th column of $\mathbf{Q}$.

For $\mathbf{F}$, the interior point method is employed to transform Eq. \eqref{eq:obj} to the unconstrained optimization problem, which can be listed as follows:
\begin{equation}
        \min_{\mathbf{F}} \left\| (\mathbf{\hat{X}}\mathbf{P})\odot \mathbf{F}\mathbf{1}_k+\mathbf{F}\mathbf{p}_0-\mathbf{y} \right\|_2^2 + \mu \sum_i\sum_j \frac{1}{F_{ij}}, 
\end{equation}
where $\mu$ is the penalty term and $\frac{1}{F_{ij}}$ is the barrier function. As a result, the gradient of $F_{ij}$ is:
\begin{equation}
        \nabla F_{ij} = 2(\langle \mathbf{\hat{x}}^i\mathbf{P}, \mathbf{f}^i \rangle + \langle \mathbf{f}^i, \mathbf{p}_0 - y_i) \rangle - \mu \frac{1}{F_{ij}^2}.
\end{equation}

For $\mathbf{p}_0$, we can obtain the closed-form solution as follows:
\begin{equation}\label{eq:updatep0}
        \mathbf{p}_0 = \mathbf{F}^{\dagger}(\mathbf{y} - (\mathbf{\hat{X}}\mathbf{P})\odot \mathbf{F}\mathbf{1}_k),
\end{equation}
where $\mathbf{F}^{\dagger}$ is the Moore-Penrose inverse of $\mathbf{F}$.

For $\mathbf{Q}$, the Lagrange multiplier method is employed to transform Eq. \eqref{eq:obj} into the unconstrained optimization problem: 
\begin{equation}\label{eq:Q}
        \min_{\mathbf{Q}}\max_{\mathbf{\varLambda}}\alpha \left\| \mathbf{X}\mathbf{Q}-\mathbf{\hat{X}} \right\|_\mathrm{F}^2 + \beta \left\| \mathbf{Q} \right\|_{2,1} + \text{tr}(\mathbf{\varLambda}(\mathbf{Q}^\mathrm{T}\mathbf{Q}-\mathbf{I})),
\end{equation}
where $\mathbf{\varLambda}$ is the Lagrange multiplier. $\mathbf{\varLambda}$ can be set as a diagonal matrix because the orthogonal constraint is imposed on $\mathbf{Q}$. The primal problem is unbounded for $\mathbf{\varLambda}$ when the orthogonal constraint holds. Therefore, the strong duality holds for Eq. \eqref{eq:Q}. We can solve the dual problem of Eq. \eqref{eq:Q} to update $\mathbf{Q}$ and $\mathbf{\varLambda}$. The dual problem is listed as follows:
\begin{equation}\label{eq:Qdual}
        \max_{\mathbf{\varLambda}}\min_{\mathbf{Q}}\alpha \left\| \mathbf{X}\mathbf{Q}-\mathbf{\hat{X}} \right\|_\mathrm{F}^2 + \beta \left\| \mathbf{Q} \right\|_{2,1} + \text{tr}(\mathbf{\varLambda}(\mathbf{Q}^\mathrm{T}\mathbf{Q}-\mathbf{I})).
\end{equation}
Only the diagonal elements of the Lagrange multiplier matrix are available when the orthogonal constraint holds. Thus, we can set $\mathbf{\varLambda} = \text{diag}(\lambda_1, \lambda_2, \cdots, \lambda_d)$. Taking the partial derivative of Eq. \eqref{eq:Qdual} w.r.t. $\mathbf{Q}$, we have:
\begin{equation}\label{eq:Qd}
        \frac{\partial \mathcal{L}}{\partial \mathbf{Q}} = 2\alpha \mathbf{X}^\mathrm{T}(\mathbf{X}\mathbf{Q}-\mathbf{\hat{X}}) + 2\beta \mathbf{Z}\mathbf{Q} + 2\mathbf{Q}\mathbf{\varLambda},
\end{equation}
where $\mathbf{Z}$ is the diagonal matrix with $Z_{ii} = \frac{1}{2\sqrt{\sum_j \mathbf{Q}_{ij}^2}}$. 

Despite $\mathbf{Z}$ dependent on $\mathbf{Q}$, we can solve Eq. \eqref{eq:Qd} by iteration. The $\mathbf{Z}$ can be easily solved when $\mathbf{Q}$ is fixed. When $\mathbf{Z}$ is fixed, let the partial derivative be zero, and then we can be transformed Eq. \eqref{eq:Qd} to the following:
\begin{equation}\label{eq:Qd1}
        (\alpha \mathbf{X}^\mathrm{T}\mathbf{X} + \beta \mathbf{Z})\mathbf{Q} + \mathbf{Q}\mathbf{\varLambda} = \alpha \mathbf{X}^\mathrm{T}\mathbf{\hat{X}}.
\end{equation}
\begin{figure*}[htbp!]
        \centering
        \subfloat[Original data]{
                \includegraphics[width=0.3\textwidth]{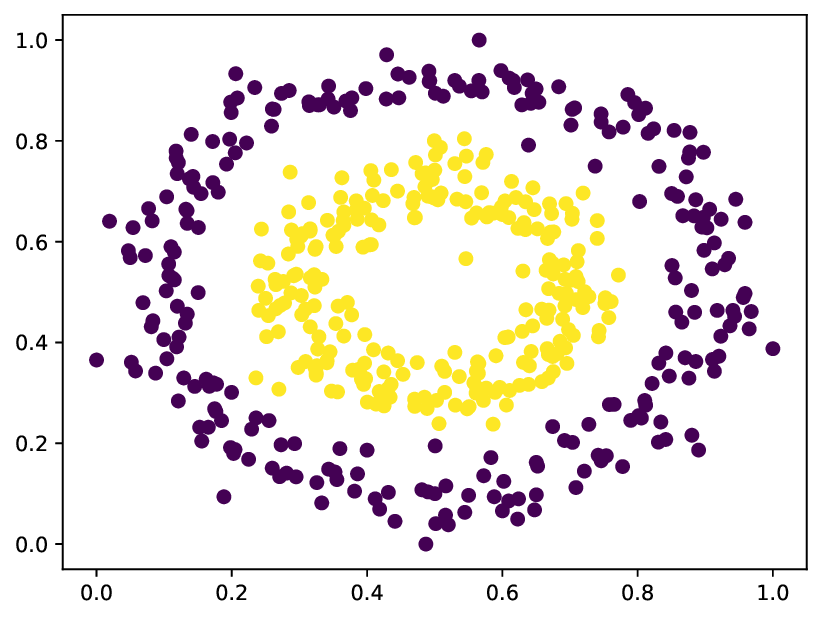}
        }
        \hfil
        \subfloat[Ours]{
                \includegraphics[width=0.3\textwidth]{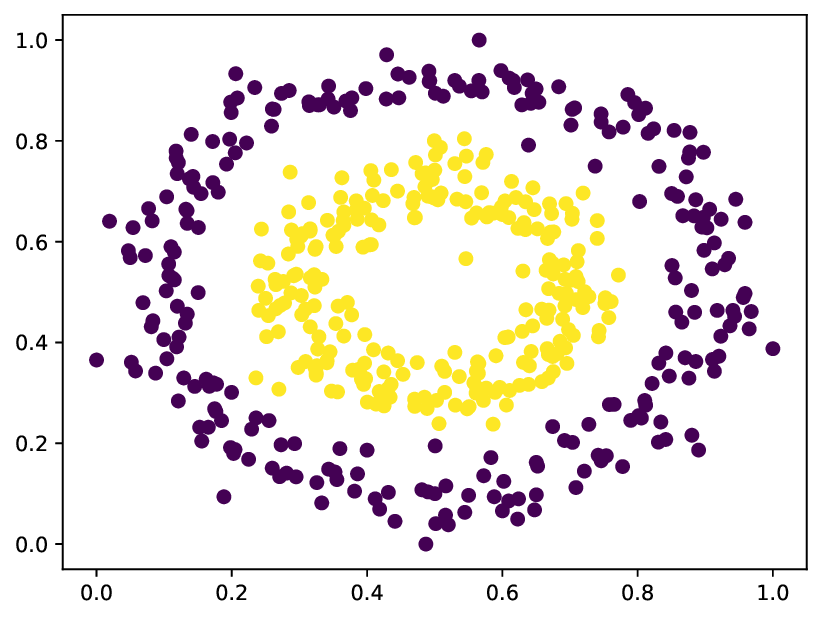}
        }
        \hfil
        \subfloat[ERFS]{
                \includegraphics[width=0.3\textwidth]{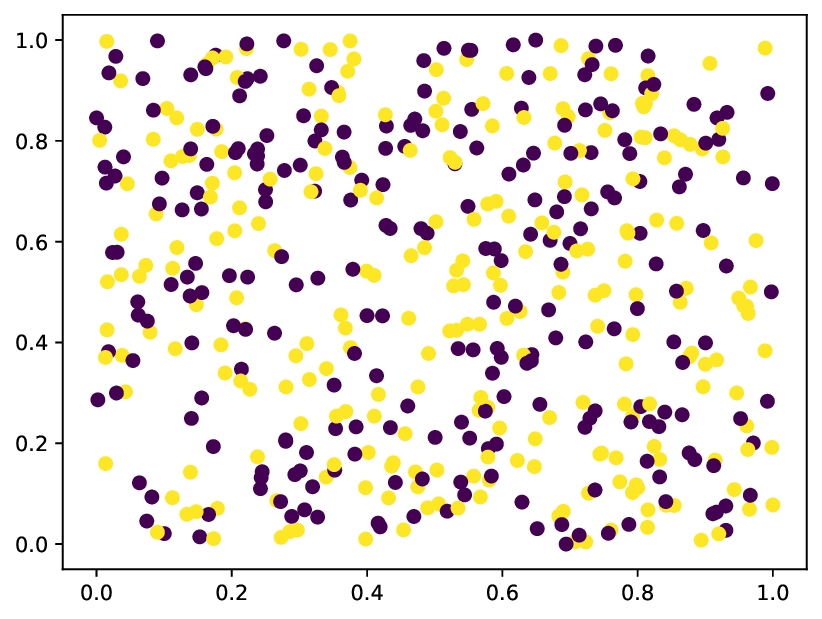}
        }
        \caption{The visualization of the selected first two features for double-circle toy data. The same color represents samples in the same class.}
        \label{fig:toy}
\end{figure*} 
Eq. \eqref{eq:Qd1} is a classical Sylvester equation that can be solved using the Bartels-Stewart algorithm. However, the Bartels-Stewart algorithm is not suitable for large-scale problems. We can leverage property of the positive semidefinite matrix to  reduce the computational complexity. Specifically, in practical application, $\mathbf{X}$ is usually normalized to the interval $[0,1]$. Therefore, $\alpha \mathbf{X}^\mathrm{T}\mathbf{X} + \beta \mathbf{Z}$ is a positive semidefinite matrix. Here, we assume that $\mathbf{\varLambda}$ is also a positive semidefinite matrix. Later, the process of making $\mathbf{\varLambda}$ a positive semi-definite matrix is elaborated.

Let $\mathbf{A} = \alpha \mathbf{X}^\mathrm{T}\mathbf{X} + \beta \mathbf{Z}$, we use SVD for $\mathbf{A}$ and $\mathbf{\varLambda}$. Then, we have:
\begin{equation}\label{eq:svd1}
        \begin{split}
                &\mathbf{A} = \mathbf{U}\mathbf{\Sigma}_1\mathbf{U}^\mathrm{T}, \\
                &\mathbf{\varLambda} = \mathbf{V}\mathbf{\Sigma}_2\mathbf{V}^\mathrm{T}.
        \end{split}        
\end{equation}
Substituting Eq. \eqref{eq:svd1} in Eq. \eqref{eq:Qd1}, we have:
\begin{equation}\label{eq:svd2}
        \mathbf{U}\mathbf{\Sigma}_1\mathbf{U}^\mathrm{T}\mathbf{Q} + \mathbf{Q}\mathbf{V}\mathbf{\Sigma}_2\mathbf{V}^\mathrm{T} = \alpha \mathbf{X}^\mathrm{T}\mathbf{\hat{X}}.
\end{equation}
Multiplying $\mathbf{U}^\mathrm{T}$ on the left and $\mathbf{V}$ on the right in Eq. \eqref{eq:svd2}, we have:
\begin{equation}
        \mathbf{\Sigma}_1\mathbf{U}^\mathrm{T}\mathbf{Q}\mathbf{V} + \mathbf{U}^\mathrm{T}\mathbf{Q}\mathbf{V}\mathbf{\Sigma}_2 = \alpha \mathbf{U}^\mathrm{T}\mathbf{X}^\mathrm{T}\mathbf{\hat{X}}\mathbf{V}.
\end{equation}
Let $\mathbf{C} = \alpha \mathbf{U}^\mathrm{T}\mathbf{X}^\mathrm{T}\mathbf{\hat{X}}\mathbf{V}$ and $\mathbf{\hat{Q}} = \mathbf{U}^\mathrm{T}\mathbf{Q}\mathbf{V}$, we have:
\begin{equation}
        \mathbf{\Sigma}_1\mathbf{\hat{Q}} + \mathbf{\hat{Q}}\mathbf{\Sigma}_2 = \mathbf{C}.
\end{equation}
Because $\mathbf{\Sigma}_1$ and $\mathbf{\Sigma}_2$ are diagonal matrices, we can obtain the solution of $\mathbf{\hat{Q}}$ as follows:
\begin{equation}
        \hat{Q}_{ij} = \frac{C_{ij}}{\Sigma_1^{ii} + \Sigma_2^{jj}}.
\end{equation}
Thus, we can obtain the solution of $\mathbf{Q}$ as follows:
\begin{equation}\label{eq:updateQ}
        \mathbf{Q} = \mathbf{U}\mathbf{\hat{Q}}\mathbf{V}^\mathrm{T}.
\end{equation}

After obtaining $\mathbf{Q}$, we can update $\mathbf{\varLambda}$ using gradient ascent. The update rule for $\mathbf{\varLambda}$ can be represented as:
\begin{equation}\label{eq:updateLambda}
        \mathbf{\varLambda}^{(t+1)} = \mathbf{\varLambda}^{(t)} + \eta (\mathbf{Q}^{\mathrm{T}}\mathbf{Q} - \mathbf{I}),
\end{equation}
where $\eta$ is the learning rate. Note that the learning rate can be adjusted to ensure that $\mathbf{\varLambda}$ is the positive semidefinite. The update algorithm for $\mathbf{Q}$ is summarized in Algorithm \ref{alg:Q}. In practice, Algorithm \ref{alg:Q} can rapidly converge to an optimal solution. Thus, we only need to execute Algorithm \ref{alg:Q} one or two times. Finally, the feature selection algorithm is summarized in Algorithm \ref{alg:obj}.
\begin{algorithm}[htbp!]
        \caption{Update $\mathbf{Q}$}
        \label{alg:Q}
        \begin{algorithmic}[1]
                \REQUIRE The training data $\mathbf{X}$ and the low-dimensional representation $\mathbf{\hat{X}}$
                \ENSURE $\mathbf{Q}$
                \STATE Initialize $\mathbf{Z}=\mathbf{I}$ and $\mathbf{\varLambda}=\mathbf{I}$;
                \REPEAT
                \STATE Update $\mathbf{Q}$ by Eq. \eqref{eq:updateQ};
                \STATE Update $\mathbf{Z}$ by $\mathbf{Z}_{ii} = \frac{1}{2\sqrt{\sum_j \mathbf{Q}_{ij}^2}}$;
                \STATE Update $\mathbf{\varLambda}$ by Eq. \eqref{eq:updateLambda};
                \UNTIL{Convergence}
        \end{algorithmic}
\end{algorithm}
\subsection{Connections to Some Related Works}
In this section, the connection between two methods, TSK-SRB~\cite{gong2024embedded} and ERFS~\cite{nie2010efficient}, and the proposed method is discussed.

TSK-SRB is a feature selection method based on TSK-FS. It transforms the form of consequent parameters to the matrix form associated with the dimension of features. Then, the group lasso loss function is used to select the significant features. The objective function of TSK-SRB is
\begin{equation}
        \min \mathcal{L}_{TSK} + \alpha \sum_{i=1}^m \left\| \mathbf{W} \right\|_{2,1},
\end{equation}
where $\mathcal{L}_{TSK}=\sum_{i=1}^n\sum_{j=1}^{c}(\hat{y}_j(x_i)-y_{i,j})^2$ is the multiple output regression loss function. The $\mathbf{W}\in \mathbb{R}^{m\times ck}$ is the matrix form of the consequent parameters, where $c$ is the number of classes. TSK-SRB employed automatic threshold segmentation to select features with high scores based on $\left\| \mathbf{w}^i \right\|_{2}$.

Our method is different from TSK-SRB in terms of the following aspects: 
\begin{enumerate}
        \item Our method selects features by learning the row-sparse projection matrix, which ensures the significance of the selected features rather than the high performance for TSK.
        \item The low-dimensional representation is employed to train TSK-FS, reducing the interference of redundant features. This can significantly improve the performance of TSK-FS while enhancing the effectiveness of the low-dimensional representation.
        \item The firing strength is directly learned, avoiding the numerical underflow.
\end{enumerate}
\begin{algorithm}[htbp!]
        \caption{Feature selection} 
        \label{alg:obj}
        \begin{algorithmic}[1]
                \REQUIRE The training data $\mathbf{X}$, the label $\mathbf{y}$ and the parameters $\alpha$, $\beta$, $\gamma$
                \STATE Initialize $\mathbf{\hat{X}}$, $\mathbf{F}$, $\mathbf{P}$, $\mathbf{Q}$, $\mathbf{p}_0$, $0<\mu<1$
                \REPEAT
                \STATE Update $\mathbf{Q}$ through Algorithm \ref{alg:Q};
                \STATE Update $\mathbf{F}$ by Eq. \eqref{eq:upf};
                \STATE Update $\mathbf{P}$ by Eq. \eqref{eq:upp};
                \STATE Update $\mathbf{p}_0$ by Eq. \eqref{eq:updatep0};
                \STATE Update $\mathbf{\hat{X}}$ by Eq. \eqref{eq:upx};
                \STATE $\mu =0.99*\mu$;
                \UNTIL{Convergence}
                \STATE Select the features by ranking the $\ell _{2,1}$-norm of $\mathbf{Q}$ in descending order.
        \end{algorithmic}
\end{algorithm}

ERFS employed the residual loss function that is not squared and $\ell _{2,1}$-norm to select the features. The objective function of ERFS is
\begin{equation}
        \min _{\mathbf{W}}\left\| \mathbf{X}^T\mathbf{W} - \mathbf{Y} \right\|_{2,1} + \alpha \left\| \mathbf{W} \right\|_{2,1}.
\end{equation}

Our method is based on TSK-FS to provide the availability of low-dimensional representation. Then, the projection matrix is utilized to find the intrinsic  subspace of data, instead of fitting the label for classification. ERFS employed linear regression to fit the label, which may run into trouble when faced with complex data.

A toy dataset is used to illustrate the difference between our method and ERFS. First, the double-circle toy data with two dimensions is generated. Then, the eight-dimensional noises with the uniform distribution are added to the data. As shown in Fig. \ref{fig:toy}, our method can select the significant first two features, while ERFS cannot.
\section{EXPERIMENTS}\label{sec:exp}
\subsection{Experimental Setup}
In this section, 18 benchmark datasets are employed to evaluate the performance of the proposed method. The datasets are listed in Table \ref{tab:1}. 
\begin{table}[htbp!]
        \centering
        \caption{Datasets Information}
        \scalebox{0.87}{\begin{tabular}{ccccc}
        \toprule
        Datasets   & \#Samples & \#Features & \#Classes & Domain           \\
        \hline
        Arrhythmia & 452       & 279        & 13        & Life             \\
        Australian & 690       & 14         & 2         & Business         \\
        Colon      & 62        & 2000       & 2         & Microarray       \\
        Control    & 600       & 60         & 6         & Synthetic        \\
        Diabetes   & 768       & 8          & 2         & Health           \\
        German     & 1000      & 20         & 2         & Financial        \\
        Isolet     & 1560      & 617        & 26        & Computer Science \\
        kr-vs-kp   & 3196      & 36         & 2         & Game             \\
        %Lung       & 203       & 3312       & 5         & Health           \\
        Madelon    & 2600      & 500        & 2         & Synthetic        \\
        %ORL        & 400       & 1024       & 40        & Image            \\
        Pima       & 768       & 8          & 2         & Health           \\
        Spambase   & 4601      & 57         & 2         & E-mail           \\
        Vote       & 435       & 16         & 2         & Social Science   \\
        Vowel      & 990       & 10         & 11        & Audio            \\
        WarpAR10P  & 130       & 2400       & 10        & Image            \\
        WarpPIE10P & 210       & 2420       & 10        & Image            \\
        Waveform   & 5000      & 40         & 3         & Physical         \\
        Wdbc       & 569       & 31         & 2         & Microarray       \\
        Yale       & 165       & 1024       & 15        & Image           \\
        \bottomrule
        \end{tabular}}
        \label{tab:1}
\end{table}
To evaluate the performance of the proposed method, we compared our method with six state-of-the-art feature selection methods (three feature selection methods based on TSK-FS and three conventional feature selection methods): DG-ALETSK~\cite{xue2023dg}, DG-TSK~\cite{xue2023double}, FRSE-TSK~\cite{xue2022adaptive}, FSOR~\cite{wu2020supervised}, MCFS~\cite{zhang2024supervised}, RJFWLF~\cite{yan2016robust}.

For our method, parameters $\alpha$, $\beta$ and $\gamma$ are tuned by the grid-search strategy from $\{0.01, 0.1, 1, 10, 100\}$. The learning rate is set as 0.01 for $\hat{\mathbf{X}}$ and $\mathbf{P}$, and the learning rate is set as 0.0001 for $\mathbf{F}$. The dimension of projection $d$ and the number of selected features are set as $\frac{m}{3}$. The same is true for the three conventional methods if they have the similar parameter to control the dimension. For other methods, we follow the settings in the original papers to search the best parameters for the best performance.

Data has been pre-processed by the min-max normalization and SVM has been employed as the classifier to evaluate the performances of all feature selection methods. In order to ensure the reliability of results, 10-fold cross validation is used. 
\begin{table*}[htbp!]
        \centering
        \caption{The classification accuracy of all methods (mean $\pm$ std)}
        \label{tab:acc}
        \begin{tabular}{cccccccc}
        \toprule
        Datasets   & DG-ALETSK~\cite{xue2023dg}        & DG-TSK~\cite{xue2023double}           & FRSE-TSK~\cite{xue2022adaptive}         & FSOR~\cite{wu2020supervised}             & MCFS~\cite{zhang2024supervised}             & RJFWLF~\cite{yan2016robust}           & Ours              \\
        \hline
        arrhythmia & $54.43\pm 2.11$  & $59.28\pm 2.81$  & $58.62\pm 2.69$  & $66.36\pm 3.66$  & $54.86\pm 1.12$  & $65.71\pm 3.02$  & $\mathbf{66.37\pm 2.15}$  \\
        australian & $85.51\pm 5.38$  & $85.51\pm 5.38$  & $85.8\pm 4.97$   & $86.09\pm 5.69$  & $57.97\pm 3.49$  & $85.8\pm 4.97$   & $\mathbf{86.96\pm 5.31}$  \\
        colon      & $79.29\pm 12.28$ & $77.62\pm 14.76$ & $72.62\pm 12.79$ & $84.05\pm 14.45$ & $75.95\pm 12.97$ & $\mathbf{88.81\pm 14.37}$ & $87.14\pm 12.38$ \\
        control    & $76.5\pm 11.94$  & $96.67\pm 1.49$  & $94.5\pm 2.24$   & $97.83\pm 1.5$   & $89.83\pm 5.29$  & $98.17\pm 2.03$  & $\mathbf{98.83\pm 1.5}$   \\
        diabetes   & $53.35\pm 10.73$ & $73.69\pm 3.97$  & $74.99\pm 2.91$  & $\mathbf{76.55\pm 3.39}$  & $60.44\pm 11.98$ & $75.12\pm 4.17$  & $76.29\pm 3.34$  \\
        german     & $57.8\pm 10.89$  & $68\pm 4.58$     & $63.2\pm 3.63$   & $68.5\pm 5.02$   & $61.1\pm 6.44$   & $68.4\pm 5.52$   & $\mathbf{76.9\pm 1.97}$   \\
        isolet     & $26.54\pm 6.33$  & $34.62\pm 4.78$  & $84.81\pm 3.05$  & $95.58\pm 2.42$  & $87.05\pm 3.08$  & $93.4\pm 1.84$   & $\mathbf{96.79\pm 1.57}$  \\
        kr-vs-kp   & $51.81\pm 5.99$  & $53.38\pm 3.17$  & $56.76\pm 1.46$  & $94.18\pm 1.51$  & $61.8\pm 4.44$   & $90.23\pm 13.19$ & $\mathbf{95.87\pm 1.22}$  \\
        madelon    & $49.73\pm 2.16$  & $58\pm 8.15$     & $52.04\pm 2.58$  & $55.58\pm 2.26$  & $51.46\pm 2.19$  & $58.73\pm 2.5$   & $\mathbf{62.69\pm 2.01}$  \\
        pima       & $51.29\pm 11.49$ & $73.18\pm 3.47$  & $76.17\pm 4.38$  & $76.3\pm 4.02$   & $57.17\pm 17.25$ & $75.39\pm 3.1$   & $\mathbf{76.56\pm 3.73}$  \\
        spambase   & $46.64\pm 9.08$  & $35.45\pm 10.77$ & $45.03\pm 17.59$ & $64.83\pm 8.64$  & $55.94\pm 4.11$  & $66.7\pm 5.21$   & $\mathbf{81.85\pm 2.09}$  \\
        vote       & $72.43\pm 13.62$ & $95.19\pm 3.86$  & $95.42\pm 3.52$  & $94.96\pm 3.78$  & $87.1\pm 5.71$   & $95.88\pm 3.18$  & $\mathbf{97.94\pm 2.37}$  \\
        vowel      & $50.1\pm 15.94$  & $58.38\pm 3.64$  & $\mathbf{88.38\pm 2.94}$  & $68.69\pm 3.67$  & $46.06\pm 11.85$ & $68.48\pm 3.66$  & $72.42\pm 2.39$  \\
        warpAR10P  & $76.15\pm 8.03$  & $43.08\pm 14.27$ & $50.77\pm 8.57$  & $80.00\pm 12.02$    & $81.54\pm 13.41$ & $\mathbf{88.46\pm 9.26}$  & $86.15\pm 8.97$  \\
        warpPIE10P & $99.52\pm 1.43$  & $22.38\pm 5.24$  & $92.38\pm 5.3$   & $99.52\pm 1.43$  & $98.57\pm 3.05$  & $\mathbf{100\pm 0}$       & $\mathbf{100\pm 0}$       \\
        waveform   & $47.34\pm 12.72$ & $\mathbf{84.98\pm 1.68}$  & $70.28\pm 3.06$  & $82.9\pm 1.39$   & $67.64\pm 4.43$  & $80.34\pm 1.77$  & $82.44\pm 2.18$  \\
        wdbc       & $71.22\pm 13.6$  & $91.92\pm 2.5$   & $96.67\pm 1.99$  & $96.31\pm 1.99$  & $89.8\pm 6.13$   & $97.02\pm 2.23$  & $\mathbf{98.07\pm 1.83}$  \\
        Yale       & $22.83\pm 9.09$  & $20.07\pm 13.05$ & $24.71\pm 8.33$  & $74.52\pm 8.03$  & $66.54\pm 10.19$ & $\mathbf{78.12\pm 8.7}$   & $75.04\pm 7.26$  \\
        \hline
        Average    & $59.58\pm 9.04$  & $62.85\pm 5.97$  & $71.28\pm 5.11$  & $81.26\pm 4.71$  & $69.49\pm 7.06$  & $81.93\pm 4.92$  & $\mathbf{84.35\pm 3.45}$  \\
        \bottomrule
        \end{tabular}
\end{table*}
\subsection{Performances Analyses for Classification}
\begin{table*}[htbp!]
        \centering
        \caption{Macro-F1 score of all methods (mean $\pm$ std)}
        \label{tab:f1}
        \begin{tabular}{cccccccc}
        \toprule
        Datasets   & DG-ALETSK~\cite{xue2023dg}        & DG-TSK~\cite{xue2023double}           & FRSE-TSK~\cite{xue2022adaptive}         & FSOR~\cite{wu2020supervised}             & MCFS~\cite{zhang2024supervised}             & RJFWLF~\cite{yan2016robust}           & Ours              \\
        \hline
        arrhythmia & $8.42\pm 4.86$   & $12.84\pm 2.78$  & $22.41\pm 7.19$  & $36.20\pm 5.5$   & $8.81\pm 2.99$   & $28.50\pm 6.04$  & $\mathbf{33.90\pm 5.17}$  \\
        australian & $85.48\pm 5.40$  & $85.48\pm 5.40$  & $85.74\pm 5.02$  & $85.96\pm 5.79$  & $55.36\pm 4.63$  & $85.74\pm 5.02$  & $\mathbf{86.88\pm 5.36}$  \\
        colon      & $71.09\pm 19.95$ & $68.17\pm 22.91$ & $60.14\pm 20.49$ & $79.73\pm 20.16$ & $67.53\pm 19.54$ & $\mathbf{85.04\pm 20.7}$  & $83.2\pm 18.59$  \\
        control    & $75.67\pm 12.97$ & $96.65\pm 1.50$  & $94.46\pm 2.27$  & $97.83\pm 1.51$  & $89.57\pm 5.57$  & $98.17\pm 2.04$  & $\mathbf{98.82\pm 1.53}$  \\
        diabetes   & $46.84\pm 6.83$  & $68.29\pm 4.64$  & $70.13\pm 3.01$  & $\mathbf{72.04\pm 4.06}$  & $55.92\pm 10.96$ & $69.5\pm 5.40$   & $71.34\pm 4.06$  \\
        german     & $44.15\pm 4.39$  & $57.11\pm 7.00$  & $52.86\pm 5.54$  & $61.64\pm 4.43$  & $52.5\pm 4.67$   & $61.64\pm 5.57$  & $\mathbf{70.47\pm 4.85}$  \\
        isolet     & $21.76\pm 7.19$  & $29.99\pm 5.29$  & $84.58\pm 3.19$  & $95.53\pm 2.47$  & $86.8\pm 3.23$   & $93.32\pm 1.90$  & $\mathbf{96.78\pm 1.59}$  \\
        kr-vs-kp   & $46.65\pm 8.36$  & $51.99\pm 3.37$  & $52.84\pm 2.07$  & $94.16\pm 1.52$  & $60.93\pm 4.54$  & $90.15\pm 13.36$ & $\mathbf{95.86\pm 1.22}$  \\
        madelon    & $45.26\pm 3.91$  & $57.61\pm 8.22$  & $51.86\pm 2.58$  & $55.08\pm 2.78$  & $51.12\pm 2.09$  & $58.61\pm 2.43$  & $\mathbf{62.53\pm 2.07}$  \\
        pima       & $46.89\pm 8.13$  & $67.11\pm 4.60$  & $71.44\pm 4.92$  & $\mathbf{71.96\pm 4.51}$  & $54.28\pm 15.75$ & $69.67\pm 3.28$  & $71.79\pm 4.46$  \\
        spambase   & $39.04\pm 10.14$ & $34.25\pm 10.09$ & $43.69\pm 17.23$ & $64.23\pm 8.36$  & $55.37\pm 4.43$  & $66.28\pm 4.74$  & $\mathbf{80.55\pm 2.33}$  \\
        vote       & $70.93\pm 14.16$ & $94.97\pm 4.06$  & $95.23\pm 3.67$  & $94.73\pm 3.97$  & $86.53\pm 5.77$  & $95.72\pm 3.30$  & $\mathbf{97.84\pm 2.51}$  \\
        vowel      & $47.22\pm 19.15$ & $58.29\pm 3.44$  & $\mathbf{88.24\pm 2.96}$  & $68.52\pm 3.51$  & $44.34\pm 12.62$ & $68.32\pm 3.56$  & $72.00\pm 2.77$  \\
        warpAR10P  & $75.07\pm 9.92$  & $35.85\pm 13.55$ & $45.57\pm 9.22$  & $77.13\pm 14.82$ & $77.07\pm 17.04$ & $\mathbf{87.07\pm 10.58}$ & $83.83\pm 10.31$ \\
        warpPIE10P & $99.6\pm 1.20$   & $16\pm 6.76$     & $92.03\pm 5.25$  & $99.6\pm 1.20$   & $98.53\pm 3.07$  & $\mathbf{100\pm 0.00}$    & $\mathbf{100\pm 0.00}$    \\
        waveform   & $45.77\pm 13.7$  & $\mathbf{84.82\pm 1.73}$  & $69.91\pm 3.04$  & $82.66\pm 1.44$  & $67.19\pm 4.29$  & $79.83\pm 1.85$  & $82.16\pm 2.26$  \\
        wdbc       & $54.8\pm 23.68$  & $91.11\pm 3.04$  & $96.41\pm 2.17$  & $95.99\pm 2.25$  & $88.73\pm 6.80$  & $96.74\pm 2.48$  & $\mathbf{97.88\pm 2.04}$  \\
        Yale       & $17.53\pm 7.42$  & $15.04\pm 11.56$ & $19.74\pm 7.08$  & $68.13\pm 10.86$ & $61.58\pm 12.46$ & $\mathbf{72.42\pm 10.49}$ & $70.16\pm 8.28$  \\
        \hline
        Average    & $52.34\pm 10.07$ & $56.97\pm 6.66$  & $66.51\pm 5.93$  & $77.84\pm 5.50$  & $64.56\pm 7.80$  & $78.15\pm 5.70$  & $\mathbf{80.88\pm 4.41}$  \\
        \bottomrule
        \end{tabular}
\end{table*}
In this section, we report the average classification accuracy and standard deviation. Table \ref{tab:acc} summarizes the classification accuracy of all methods in percentage. The best results are highlighted in bold. Based on Table \ref{tab:acc}, we have the following observations. (1) Compared with methods based on TSK-FS, our method outperforms they on almost all datasets by a significant margin. This is because they leverage the gate function to control the consequent or antecedent parameters for training TSK-FS. In other words, even though the gate function can select the most important features to fit the label, different parameters are used in the fitting process simultaneously, resulting in the selected features not having excellent discriminability. Our method adds an additional term to select significant features. Meanwhile, TSK-FS provides the availability of the low-dimensional representation, leading to the outstanding discriminability of the selected features. (2) FSOR and RJFWLF are comparable to our method in terms of classification accuracy on most datasets. For complex datasets German, Madelon and Spambase, the selected features by our method have better discriminability than FSOR and RJFWLF, leading to the significant performance difference. While FSOR and RJFWLF are based on the linear regression model, which cannot captur the intrinsic subspace of complex data. (3) It is clear that our method achieves the best results on most datasets, and the average classification accuracy is 84.35\%, which is 2.42\% higher than the second-best method RJFWLF. The part of reason is that our method applies the low-dimensional representation to train TSK-FS, which can reduce the interference of redundant features and fit more complex data, enhancing the performance of feature selection significantly. 

In order to further evaluate the performance of the selected features, Macro-F1 score has been employed to measure the performance of the selected features, which is more suitable for real-world scenarios, such as disease diagnosis. By observing table \ref{tab:f1}, we can obtain that our method also outperforms other methods in terms of Macro-F1 score, similar to the classification accuracy. In other words, our method take into account both precision and recall, which can be better implemented in practical applications. All in all, our method can obtain the best performance in terms of classification accuracy and Macro-F1 score on most datasets, which demonstrates the effectiveness of the proposed method.
\subsection{Face Recognition}
\begin{figure*}[htbp!]
        \centering
        \captionsetup[subfloat]{labelsep=none,format=plain,labelformat=empty,font={scriptsize},textfont=normalfont}
        \subfloat{
                \includegraphics[width=0.1\textwidth]{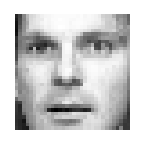}
        }
        \subfloat{
                \includegraphics[width=0.1\textwidth]{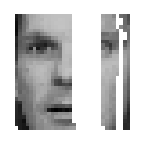}
        }
        \subfloat{
                \includegraphics[width=0.1\textwidth]{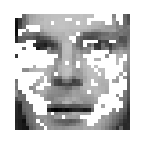}
        }
        \subfloat{
                \includegraphics[width=0.1\textwidth]{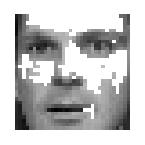}
        }
        \subfloat{
                \includegraphics[width=0.1\textwidth]{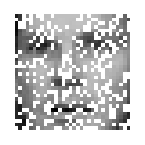}
        }
        \subfloat{
                \includegraphics[width=0.1\textwidth]{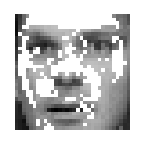}
        }
        \subfloat{
                \includegraphics[width=0.1\textwidth]{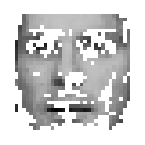}
        }
        \subfloat{
                \includegraphics[width=0.1\textwidth]{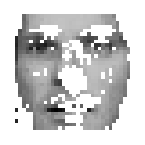}
        }

        \subfloat{
                \includegraphics[width=0.1\textwidth]{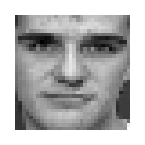}
        }
        \subfloat{
                \includegraphics[width=0.1\textwidth]{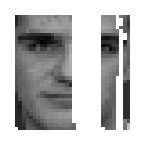}
        }
        \subfloat{
                \includegraphics[width=0.1\textwidth]{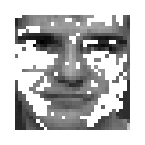}
        }
        \subfloat{
                \includegraphics[width=0.1\textwidth]{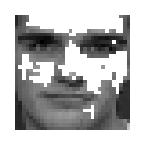}
        }
        \subfloat{
                \includegraphics[width=0.1\textwidth]{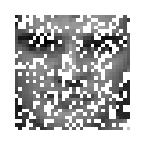}
        }
        \subfloat{
                \includegraphics[width=0.1\textwidth]{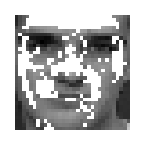}
        }
        \subfloat{
                \includegraphics[width=0.1\textwidth]{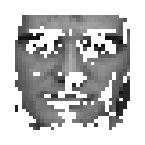}
        }
        \subfloat{
                \includegraphics[width=0.1\textwidth]{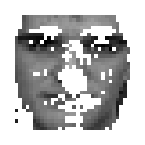}
        }

        \subfloat{
                \includegraphics[width=0.1\textwidth]{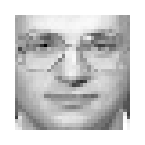}
        }
        \subfloat{
                \includegraphics[width=0.1\textwidth]{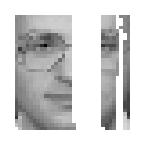}
        }
        \subfloat{
                \includegraphics[width=0.1\textwidth]{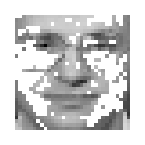}
        }
        \subfloat{
                \includegraphics[width=0.1\textwidth]{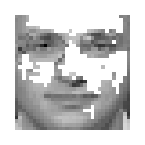}
        }
        \subfloat{
                \includegraphics[width=0.1\textwidth]{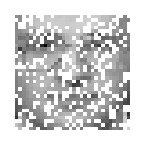}
        }
        \subfloat{
                \includegraphics[width=0.1\textwidth]{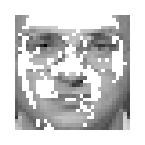}
        }
        \subfloat{
                \includegraphics[width=0.1\textwidth]{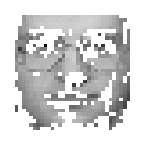}
        }
        \subfloat{
                \includegraphics[width=0.1\textwidth]{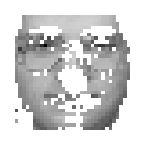}
        }

        \subfloat{
                \includegraphics[width=0.1\textwidth]{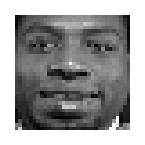}
        }
        \subfloat{
                \includegraphics[width=0.1\textwidth]{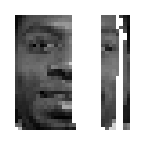}
        }
        \subfloat{
                \includegraphics[width=0.1\textwidth]{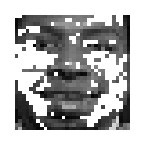}
        }
        \subfloat{
                \includegraphics[width=0.1\textwidth]{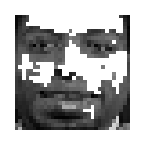}
        }
        \subfloat{
                \includegraphics[width=0.1\textwidth]{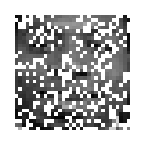}
        }
        \subfloat{
                \includegraphics[width=0.1\textwidth]{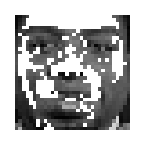}
        }
        \subfloat{
                \includegraphics[width=0.1\textwidth]{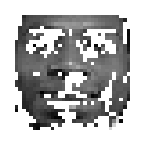}
        }
        \subfloat{
                \includegraphics[width=0.1\textwidth]{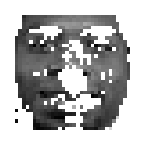}
        }

        \subfloat[(a)]{
                \includegraphics[width=0.1\textwidth]{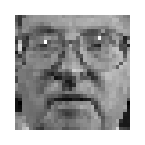}
        }
        \subfloat[(b)]{
                \includegraphics[width=0.1\textwidth]{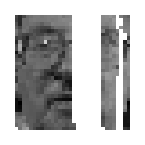}
        }
        \subfloat[(c)]{
                \includegraphics[width=0.1\textwidth]{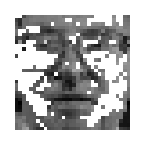}
        }
        \subfloat[(d)]{
                \includegraphics[width=0.1\textwidth]{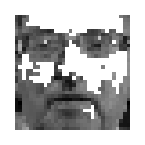}
        }
        \subfloat[(e)]{
                \includegraphics[width=0.1\textwidth]{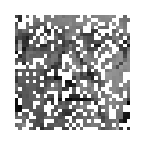}
        }
        \subfloat[(f)]{
                \includegraphics[width=0.1\textwidth]{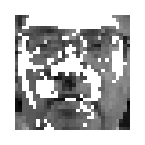}
        }
        \subfloat[(g)]{
                \includegraphics[width=0.1\textwidth]{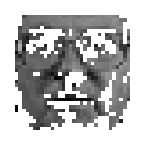}
        }
        \subfloat[(h)]{
                \includegraphics[width=0.1\textwidth]{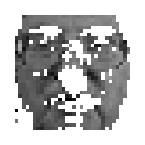}
        }
        \caption{The selected features are marked in white based on (a) original image, (b) DG-ALETSK, (c) DG-TSK, (d) FRSE-TSK, (e) FSOR, (f) MCFS, (g) RJFWLF and (h) our method.}
        \label{fig:face}
\end{figure*}

In order to visually observe the selected features, the ORL dataset\footnote{http://www.cad.zju.edu.cn/home/dengcai/Data/FaceData.html} has been employed to evaluate the performance of the proposed method. The ORL dataset contains 400 images of 40 subjects with 10 images per subject. Each image is resized to 32$\times$32 pixels. In this experiment, 50\% of samples has been randomly extracted for training. The top 30\% of features, the pixels of image, have been selected by our method and other methods. The selected pixels have been marked in white in the image. As shown in Fig. \ref{fig:face}, conclusions are summarized as follows. (1) The selected features by TSK-FS based methods are centralized. They select the features by ranking the $\ell _{2}$-norm of consequent parameters, resulting in the secondary facial features being selected. (2) FSOR utilizes the orthogonal regression to select features. Thus, it selects features that roughly cover the image to fit data. (3) MCFS considers multi-center to choice nearest samples after projection, which selects more concentrated pixels. (4) RJFWLF captures some significant features, benefiting by $\ell _{2,1}$-norm being applied to linear regression, but it cannot capture the feature of nose. For our method, the most significant features are caught, such as eyebrows, eyes, nose, mouth and face contour. These features are most discriminability for face recognition.
\renewcommand{\floatpagefraction}{.9}
\subsection{Statistical Analysis}
To further analyze the performance of all methods on all datasets, we have conducted the Friedman test~\cite{friedman1940comparison} and the Bonferroni-Dunn test~\cite{dunn1961multiple}. The definition of Chi-Square is listed as follows:
\begin{equation}
        \chi ^{2} = \frac{12N}{k(k+1)}\left( \sum _{i=1}^{k}R_{i}^{2} - \frac{k(k+1)^{2}}{4} \right),
\end{equation}
\begin{figure}[htbp!]
        \centering
        \includegraphics[width=0.5\textwidth]{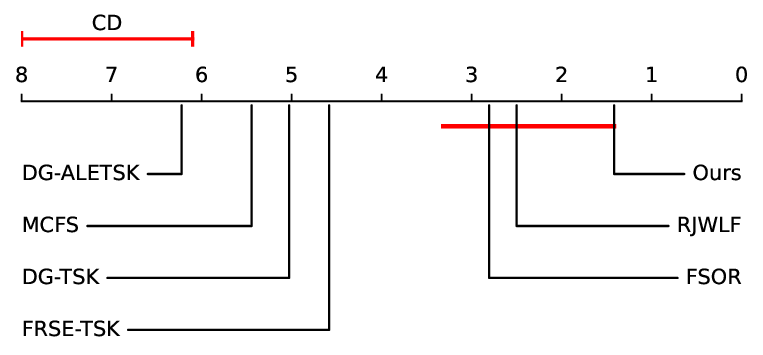}
        \caption{Comparison of our method against competitors with the Bonferroni-Dunn test. Our method differs significantly from the method that is not connected to our method in performance.}
        \label{fig:CD}
\end{figure}
where $k$ is the number of datasets, $N$ is the number of methods, and $R_{i}$ denotes the average rank of $i$-th method on all datasets. Then, the Friedman statistical formula is defined as follows:
\begin{equation}
        F = \frac{(N-1)\chi ^{2}}{N(k-1)-\chi ^{2}},
\end{equation}
where $F$ follows an F-distribution with $k-1$ and $N-1$ degrees of freedom. Then, we can obtain the Friedman statistic $F=34.5799$ and the critical value is 2.1887.

Let the null hypothesis be that the performance of all algorithms is equivalent. The null hypothesis is rejected since the Friedman statistic is greater than the critical value. Thus, we can conclude that the performance of all methods is significantly different. To further analyze the difference between the performance of all methods, the post-hoc Bonferroni-Dunn test has been conducted. The critical difference (CD) is calculated as follows:
\begin{equation}
        CD = q_{\alpha}\sqrt{\frac{k(k+1)}{6N}},
\end{equation}
where $\alpha=0.05$ is the significance level, and $q_{\alpha}=2.638$. Then, we can obtain $CD=1.8995$. From Fig. \ref{fig:CD}, we can observe that our method is significantly better than other methods, except for RJFWLF and FSOR. The average ranking of our method is approximately one position higher than the second. 
\subsection{Parameter Sensitivity Analysis}
\begin{figure*}[htbp!]
        \centering
        \captionsetup[subfloat]{labelsep=none,format=plain,labelformat=empty,font={scriptsize},textfont=normalfont}
        \subfloat[\enspace \enspace \enspace (a) Arrhythmia]{
                \includegraphics[width=0.23\textwidth]{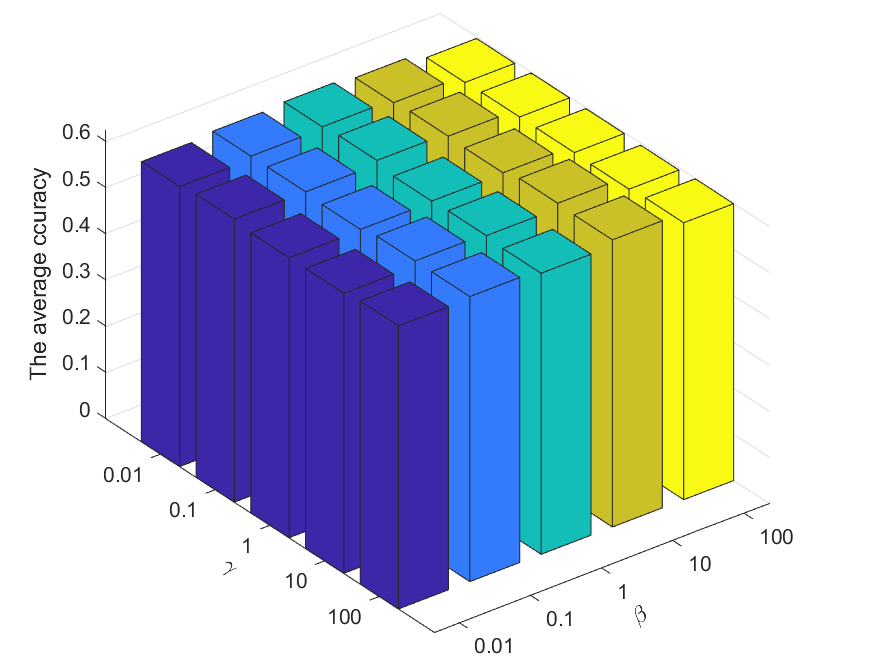}
        }
        \subfloat[\enspace \enspace \enspace (a) Control]{
                \includegraphics[width=0.23\textwidth]{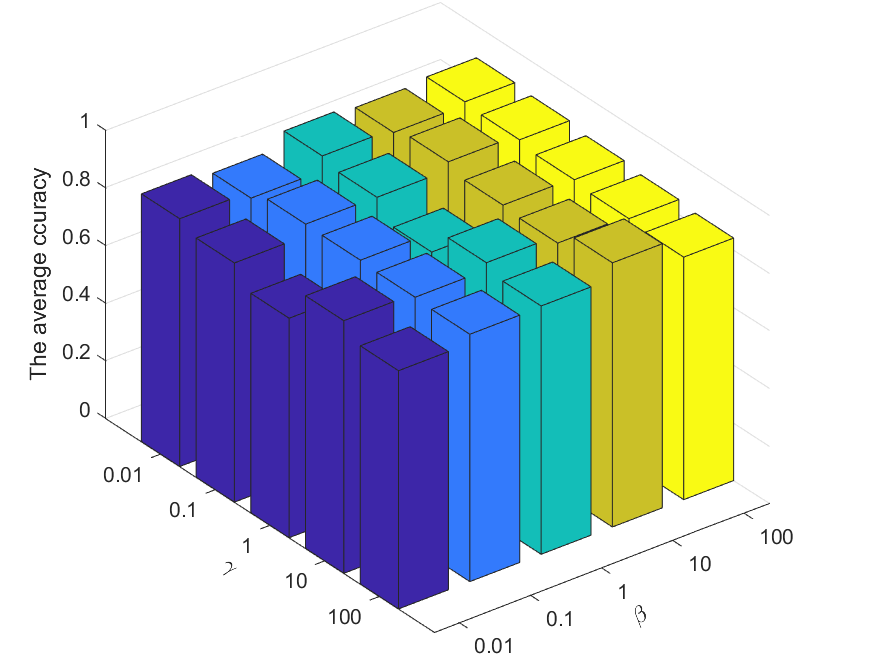}
        }
        \subfloat[\enspace \enspace \enspace (a) Vote]{
                \includegraphics[width=0.23\textwidth]{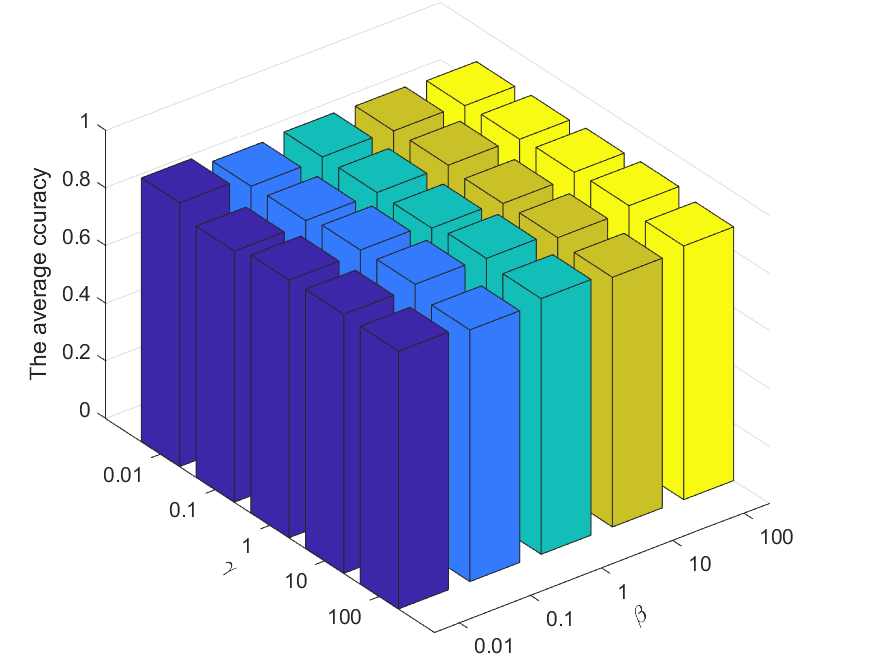}
        }
        \subfloat[\enspace \enspace \enspace (a) Yale]{
                \includegraphics[width=0.23\textwidth]{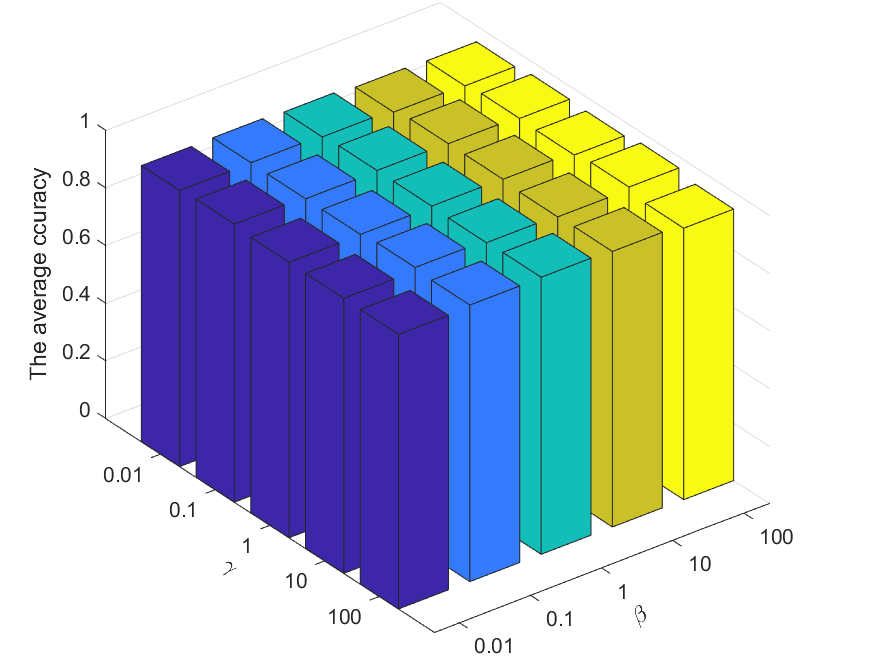}
        }

        \subfloat[\enspace \enspace \enspace (b) Arrhythmia]{
                \includegraphics[width=0.23\textwidth]{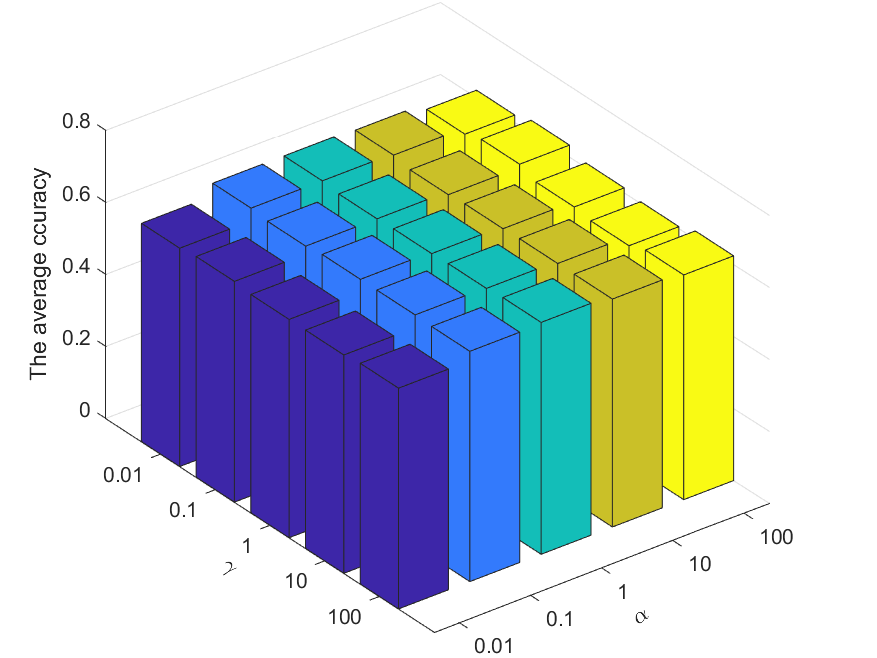}
        }
        \subfloat[\enspace \enspace \enspace (b) Control]{
                \includegraphics[width=0.23\textwidth]{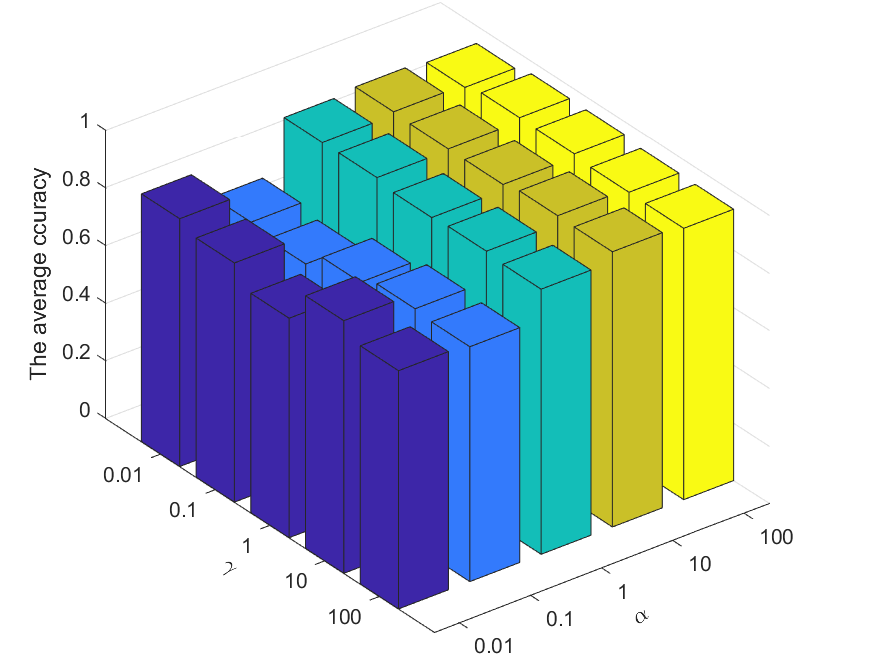}
        }
        \subfloat[\enspace \enspace \enspace (b) Vote]{
                \includegraphics[width=0.23\textwidth]{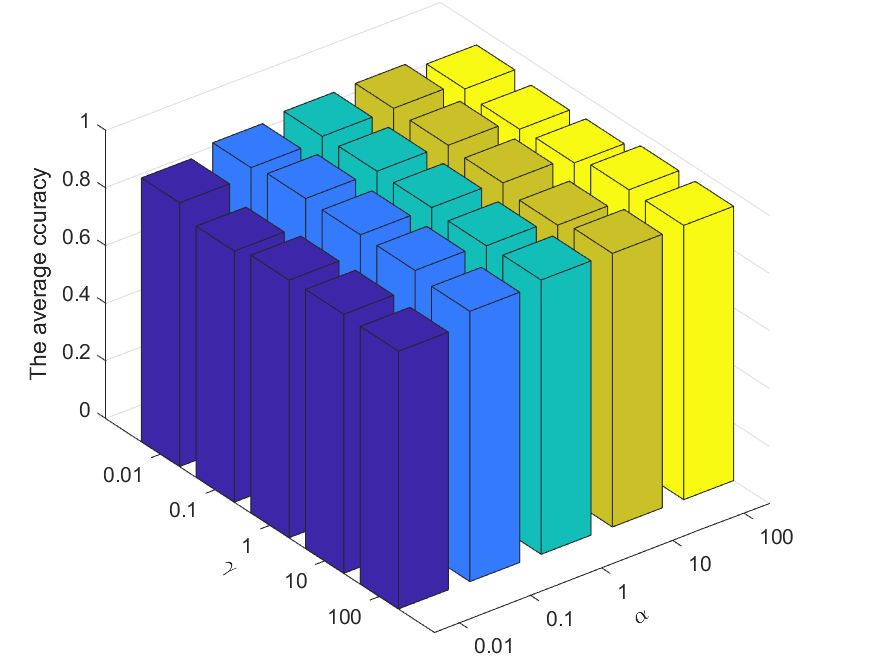}
        }
        \subfloat[\enspace \enspace \enspace (b) Yale]{
                \includegraphics[width=0.23\textwidth]{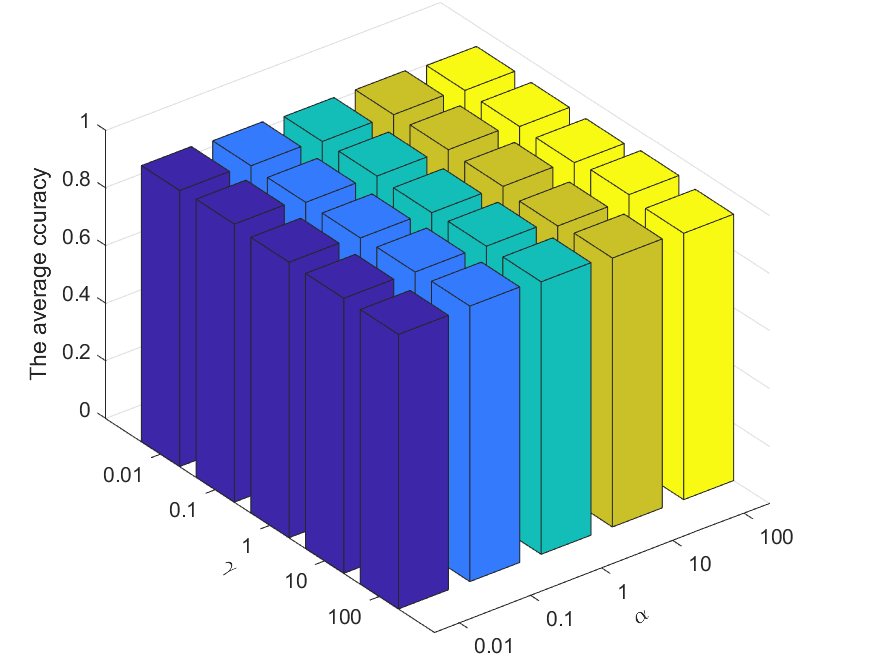}
        }

        \subfloat[\enspace \enspace \enspace (c) Arrhythmia]{
                \includegraphics[width=0.23\textwidth]{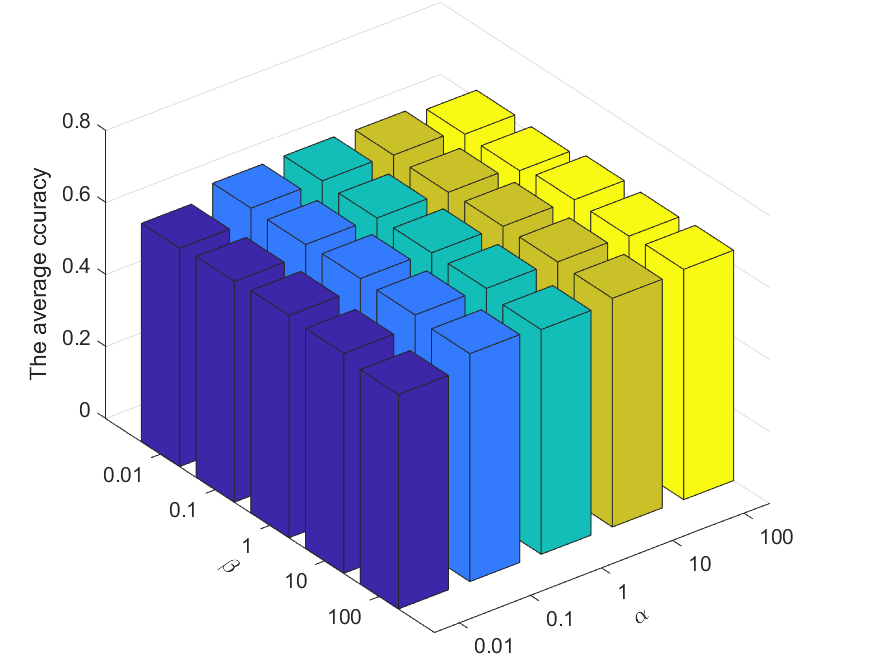}
        }
        \subfloat[\enspace \enspace \enspace (c) Control]{
                \includegraphics[width=0.23\textwidth]{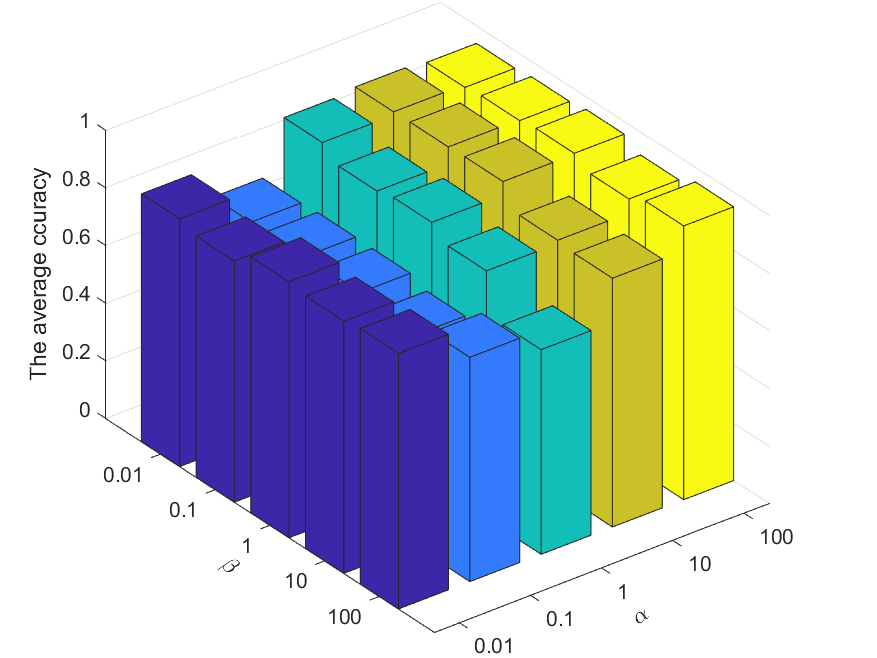}
        }
        \subfloat[\enspace \enspace \enspace (c) Vote]{
                \includegraphics[width=0.23\textwidth]{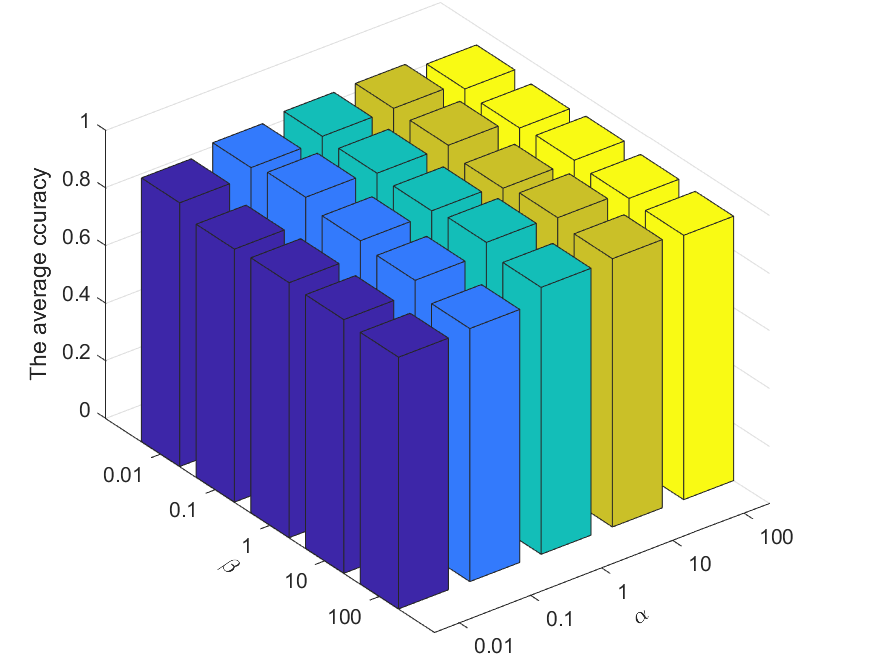}
        }
        \subfloat[\enspace \enspace \enspace (c) Yale]{
                \includegraphics[width=0.23\textwidth]{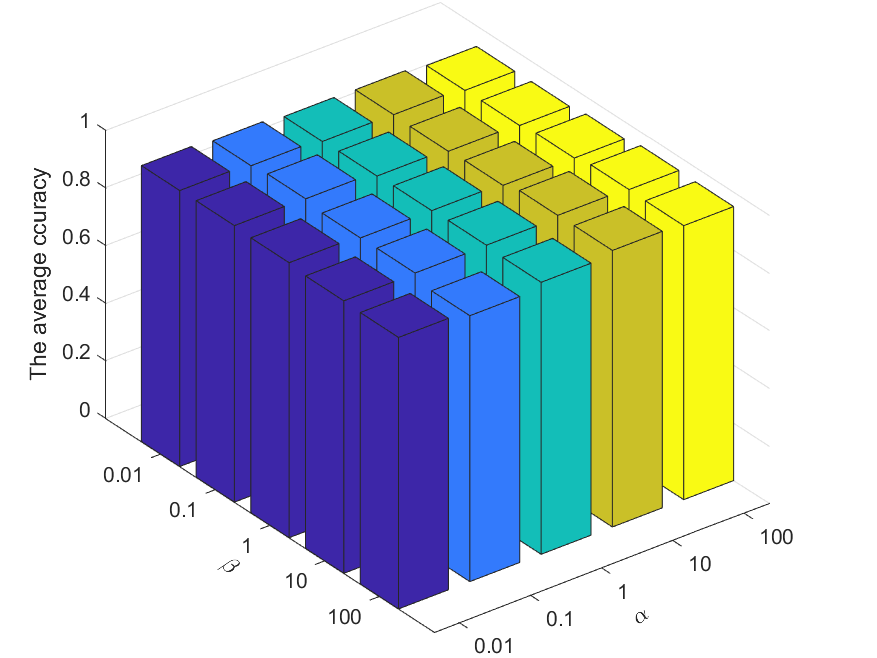}
        }
        \caption{The classification accuracy of our method with different parameters on arrhythmia, control, vote and Yale datasets. The parameters $\alpha$, $\beta$, and $\gamma$ are fixed in (a), (b), and (c) with the remaining two parameters tuned, respectively.}
        \label{fig:para}
\end{figure*}

In our method, three parameters are involved, $\alpha$, $\beta$ and $\gamma$. Due to limited length, we conducted the parameter sensitivity analysis on four datasets, Arrhythmia, Control, Vote and Yale. From Fig. \ref{fig:para}, we can observe that our method is insensitive to parameters in most cases when one parameter is fixed and the remaining parameters are tuned. Moreover, on Control dataset, the classification accuracy is relatively low when $\alpha$ is smaller than 1. The phenomenon might be caused by the fact that the constraint on subspace learning is weak, resulting in the poor discrimination of the selected features.
\renewcommand{\floatpagefraction}{.9}
\section{Conclusion}\label{sec:con}
This paper proposed a feature selection method based on TSK-FS and subspace learning. The low-dimensional representation has been employed to train TSK-FS, which can reduce the interference of redundant features and enhance the performance of TSK-FS for classification. Due to the low-dimensional representation is unknown, the assumption that there exists a low-dimensional intrinsic representation w.r.t. the original data has been utilized to project the original data into a low-dimensional subspace. Both of these can largely ensure that the learned low-dimensional representation is accurate. Then, the $\ell _{2,1}$-norm has been employed to select most discriminative features. The experimental results on 18 datasets have demonstrated that our method can achieve the best performance in terms of classification accuracy and Macro-F1 score in most cases. However, since the firing strength w.r.t. the test set is unknown, our method cannot be utilized as a classifier. In other words, TSK-FS in our method does not have the ability to predict labels of the test set. In the future, we will focus on the generalization of TSK-FS to accurately predict labels.
% \section*{Acknowledgments}
% *
% {\appendix[pass]
% pass}

\bibliography{IEEEabrv,main.bib}
\bibliographystyle{IEEEtran}

\end{document}